\documentclass[10pt,twocolumn,letterpaper]{article}

\usepackage{wacv}
\usepackage{times}
\usepackage{epsfig}
\usepackage{graphicx}
\usepackage{amsmath}
\usepackage{amssymb}

\usepackage{adjustbox}
\usepackage{caption}
\usepackage{subcaption}
\usepackage{enumitem}
 \usepackage[normalem]{ulem} 
\usepackage[table, svgnames, dvipsnames]{xcolor}
\usepackage{makecell, cellspace, caption}
\usepackage{pifont}
\usepackage[dvipsnames]{xcolor}
\usepackage{algorithm}
\usepackage{algpseudocode}

\def\wacvPaperID{0294} 

\wacvfinalcopy 

\ifwacvfinal
\fi

\ifwacvfinal
\usepackage[breaklinks=true,breaklinks=true,colorlinks,bookmarks=false]{hyperref}
\else
\usepackage[pagebackref=true,breaklinks=true,colorlinks,bookmarks=false]{hyperref}
\fi

\DeclareUnicodeCharacter{2212}{-} 
\begin{document}

\title{Multi-Domain Incremental Learning for Semantic Segmentation}

\author{Prachi Garg$^1$ \and Rohit Saluja$^1$ \and Vineeth N Balasubramanian$^2$ \and Chetan Arora$^3$ \and
Anbumani Subramanian$^1$ \hspace{10mm} C.V. Jawahar$^1$\\
$^1$CVIT - IIIT Hyderabad, India \hspace{5mm} $^2$IIT Hyderabad, India \hspace{5mm} $^3$IIT Delhi, India\\
{\tt \small $^1$prachigarg2398@gmail.com,~$^1$rohit.saluja@research.iiit.ac.in, $^2$vineethnb@iith.ac.in,}\\ {\tt \small $^3$chetan@cse.iitd.ac.in, $^1$\{anbumani, jawahar\}@iiit.ac.in}
}

\maketitle

\ifwacvfinal
\thispagestyle{empty}
\fi

\begin{abstract}
Recent efforts in multi-domain learning for semantic segmentation attempt to learn multiple geographical datasets in a universal, joint model. A simple fine-tuning experiment performed sequentially on three popular road scene segmentation datasets demonstrates that existing segmentation frameworks fail at incrementally learning on a series of visually disparate geographical domains. When learning a new domain, the model catastrophically forgets previously learned knowledge. In this work, we pose the problem of multi-domain incremental learning for semantic segmentation. Given a model trained on a particular geographical domain, the goal is to (i) incrementally learn a new geographical domain, (ii) while retaining performance on the old domain, (iii) given that the previous domain's dataset is not accessible. We propose a dynamic architecture that assigns universally shared, domain-invariant parameters to capture homogeneous semantic features present in all domains, while dedicated domain-specific parameters learn the statistics of each domain. Our novel optimization strategy helps achieve a good balance between retention of old knowledge (stability) and acquiring new knowledge (plasticity). We demonstrate the effectiveness of our proposed solution on domain incremental settings pertaining to real-world driving scenes from roads of Germany (Cityscapes), the United States (BDD100k), and India (IDD). \footnote{Code is available at \\                    \url{https://github.com/prachigarg23/MDIL-SS}}
\end{abstract}


\begin{figure}[ht]
\begin{center}
\includegraphics[width=\linewidth]{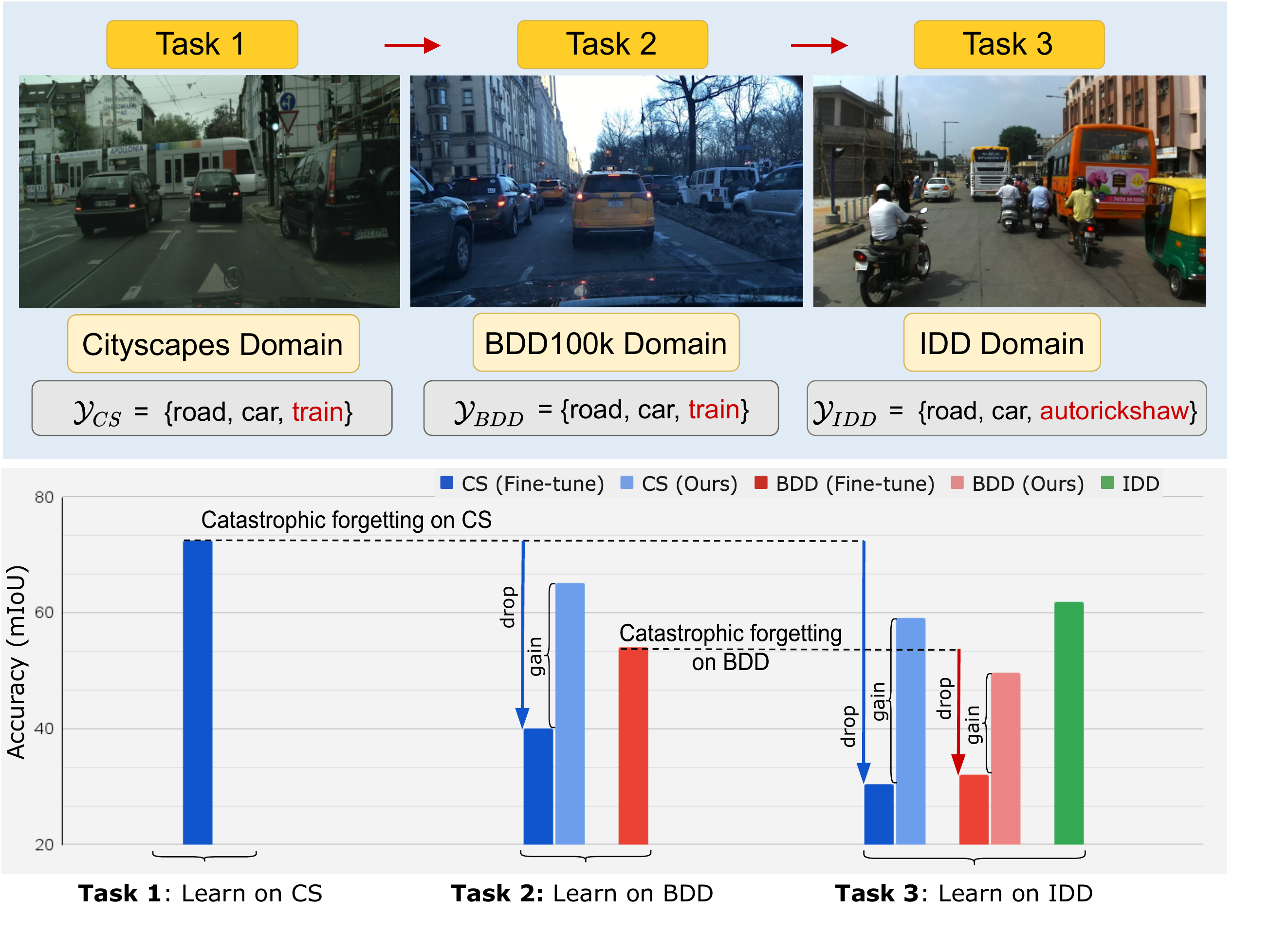} 
\end{center}
\vspace{-5mm}
\caption{\textit{Top row:} Our setting over three incremental tasks: learning a model on CS (task 1), followed by learning on BDD (task 2) and IDD (task 3). The domains have non-overlapping label spaces, where the black categories are shared among all domains, and red categories are domain-specific. \textit{Bottom row:} The problem of \textbf{catastrophic forgetting}: as the CS model is fine-tuned on BDD and further on IDD, we witness a sharp degradation in performance of old datasets; our method significantly mitigates this forgetting.}
\vspace{-8pt}
\label{fig:problem statement}
\end{figure}


\vspace{-10pt}
\section{Introduction}
\vspace{-5pt}
Driving is a skill that humans do not forget under natural circumstances. They can easily drive in multiple geographies. This shows that humans are naturally capable of lifelong learning and barely forget previously learned visual patterns when faced with a domain shift or given new objects to identify. In recent times, there has been an active interest in developing universal vision systems capable of performing well in multiple visual domains. We ask the question: can a semantic segmentation model trained on road scenes of a particular city extend to learn novel geographic environments? 

Consider learning incrementally over three autonomous driving datasets: Cityscapes $\rightarrow$ BDD100k $\rightarrow$ Indian Driving Dataset. We conduct a fine-tuning experiment over these three datasets and find that deep neural networks forget previously acquired knowledge when trained on novel geographic domains. The degradation of performance is evident in Figure~\ref{fig:problem statement}. This phenomenon where new knowledge overwrites previous knowledge is referred to as catastrophic forgetting~\cite{mccloskey1989catastrophic} in incremental learning. We observe that when shifting from one geography to another, catastrophic forgetting can be due to two factors: (i) a domain shift is encountered in the road scene environment due to varying background conditions, such as driving culture, illumination, and weather; and (ii) the label space for semantic segmentation might change when encountered with novel classes in the new geography while missing some of the old classes. 

Most semantic segmentation research today focuses on developing models that are specialized to a specific dataset or environment. They fail to work in continual learning settings, where we want to extend the scope of our autonomous driving model to road environments with a potential domain shift. In the absence of an incremental model, a naive way to solve this problem is to train a separate, independent model for each geography, store all models and deploy the corresponding model when the road scene environment changes. Another option is to store data from all these domains and re-train a single, joint model from scratch each time a new domain's data is collected. Both these approaches involve a significant computational overhead, are not scalable, and data-inefficient as it requires storing large amounts of data that may be proprietary or unavailable. Moreover, when training separate models, the new domain cannot benefit from an old one's existing knowledge (forward transfer in continual learning \cite{delange2021continual}).

Kalluri~\etal~\cite{kalluri2019universal} proposed a universal semi-supervised semantic segmentation technique that models multiple geographic domains simultaneously in a universal model. 
Their method requires simultaneous access to all the datasets involved and does not follow the incremental learning setting. We extend this literature by considering the case of incremental learning where multiple domains are learnt in a single model, \textit{sequentially}, eliminating the need to store previously learnt data. We draw inspiration from existing literature on multi-domain incremental learning (MDIL) for classification~\cite{rebuffi2017learning,rebuffi2018efficient,mallya2018piggyback}, and reparametrize a semantic segmentation architecture into domain-invariant parameters (shared among domains) and domain-specific parameters that are exclusively added, trained on and used for each novel domain being learnt. To the best of our knowledge, our work is the first attempt at MDIL for semantic segmentation. Our key contributions can be outlined as follows: 

\vspace{-7pt}
\begin{enumerate}[leftmargin=*]
\setlength\itemsep{-0.1em}
    \item We define the problem of multi-domain incremental semantic segmentation and propose a dynamic framework that reparameterizes the network into a set of domain-invariant and domain-specific parameters. We achieve this with a 78.83\% parameter sharing across all domains. 
    \item In continual learning, plasticity is the ability to acquire new knowledge, while stability refers to retaining existing knowledge~\cite{mermillod2013stability}. Our primary objective in this work is to tackle this stability-plasticity dilemma. We propose a novel optimization strategy designed to fine-tune the domain-invariant and domain-specific layers differently towards a good stability-plasticity trade-off. In a first, we find a combination of differential learning rates and domain adaptive knowledge distillation to be highly effective towards achieving this goal. 
    \item We consider the challenging issue of non-overlapping label spaces in multi-domain incremental semantic segmentation owing to its relevance in real-world autonomous driving scenarios. We show that our model performs well on both: (i) datasets that have a domain shift but an overlapping label space (Cityscapes $\rightarrow$ BDD100k); (ii) datasets that have non-overlapping label spaces in addition to domain shift (Cityscapes $\rightarrow$ Indian Driving Dataset). We also analyze forward transfer and domain interference in these cases (Section \ref{comparisons on 2 scenarios}). 
\end{enumerate}

\begin {table*}[ht]
\centering
\begin{center}
\resizebox{\textwidth}{!}{%
    \begin {tabular} {|p{3.3cm}|c|c|c|c|c|c|c|c|c|c|c}
    \hline
    Problem Setting & Sequential & \multicolumn{2}{c|}{Differences, Source vs target} & \multicolumn{2}{c|}{Data (availability, supervision)} & Goals & \multicolumn{2}{c|}{Solution Type} \\
     &  & Label Space & Domain Shift & \quad Source  \quad & Target & & Task-Aware & Multi-Head\\
    \hline
    UDA~\cite{vu2019advent, wang2020classes, yang2020fda} & \checkmark & same & \checkmark & \checkmark & \checkmark (unlabeled) & learn new & $\times$ & $\times$ \\
    Class-IL~\cite{michieli2019incremental, cermelli2020modeling, douillard2020plop, michieli2021continual} & \checkmark & different & $\times$ & $\times$ & \checkmark & retain old, learn new & $\times$ & $\times$ \\
    MDL~\cite{kalluri2019universal} & $\times$ & different & \checkmark & \checkmark & \checkmark & retain all & \checkmark & \checkmark \\
    \hline
    \textbf{MDIL} \textit{(ours)} & \checkmark & different & \checkmark & $\times$ & \checkmark & retain old, learn new & \checkmark & \checkmark \\ 
    \hline 
\end{tabular}%
}
\end{center}
\vspace{-11pt} 
\caption{A comparison of different semantic segmentation settings: Unsupervised Domain Adaptation (UDA), Class Incremental Learning (Class-IL), Multi-Domain Learning (MDL) and Multi-Domain Incremental Learning (MDIL)}
\label{Tab: problem setting}
\end{table*}

\vspace{-14pt}
\section{Related Work}
\vspace{-4pt}
\subsection{Incremental Learning}
\vspace{-5pt}
Incremental learning (IL) involves lifelong learning of new concepts in an existing model without forgetting previously learned concepts. IL in computer vision has most widely been studied for image classification~\cite{parisi2019continual,de2019continual}, where the methods can be broadly grouped into three categories~\cite{de2019continual}: memory or replay-based, regularization-based, and parameter isolation-based methods. Replay-based techniques store previous experience either implicitly via generative replay~\cite{shin2017continual,wu2018memory,ostapenko2019learning} or explicitly~\cite{rebuffi2017icarl,castro2018end,hou2019learning,wu2019large} in the form of raw samples or dataset statistics of previous data. Regularization-based methods can be further categorized as prior-focused~\cite{zenke2017continual,kirkpatrick2017overcoming,chaudhry2018riemannian,aljundi2018memory} and data-focused methods~\cite{hinton2015distilling,li2017learning,dhar2019learning}. In parameter isolation methods~\cite{mallya2018packnet, mallya2018piggyback, rebuffi2018efficient, aljundi2017expert}, additional task-specific parameters are added to a dynamic architecture for each new task. 

\vspace{3pt}
\noindent \textbf{Multi-Domain Incremental Learning.}
Multi-domain IL is concerned with sequentially learning a single task, say image classification, on multiple visual domains with possibly different label spaces. The earliest works in this space on the classification task are Progressive Neural Networks~\cite{rusu2016progressive}, Dynamically Expandable Networks (DENs)~\cite{yoon2017lifelong}, and attaching controller modules to a base network~\cite{rosenfeld2018incremental}. \cite{mallya2018piggyback} and \cite{mancini2018adding} learn a domain-specific binary mask over a fixed backbone architecture to get a compact and memory-efficient solution. Other works using parameter-isolation based techniques dedicate a domain-specific subset of parameters to each unique task, to mitigate forgetting by construction.
To this end, Rebuffi~\etal introduced residual adapters in series~\cite{rebuffi2017learning} and parallel~\cite{rebuffi2018efficient} in an attempt to define universal parametrizations for multi-domain networks by using certain domain-specific and shared network parameters. Other recent works~\cite{guo2019depthwise,bulat2020incremental,ebrahimi2020adversarial,singh2020calibrating} also share a similar approach, but focus on the classification task.  
Recently,~\cite{liu2020multi} proposed incremental learning across various domains and categories for object detection. Also related to our work is \textit{multi-task} incremental learning~\cite{kanakis2020reparameterizing} over tasks like edge detection and human parts segmentation. 
Our work in multi-domain incremental learning, while inspired by these methods, seeks to address the semantic segmentation setting for the first time.

\vspace{3pt} \noindent \textbf{Incremental Learning for Semantic Segmentation.}
IL methods have been developed for semantic segmentation in recent years, although from a \textbf{\textit{class-incremental}} perspective.
\cite{michieli2019incremental,cermelli2020modeling} and~\cite{klingner2020class} were the first to solve class-IL for semantic segmentation.
Recent methods~\cite{cermelli2020modeling, douillard2020plop, michieli2021continual} focus on the problem of semantic shift in the background class distribution, which is typical to \textit{strict} class incremental learning for semantic segmentation. In their setting, labels occurring in previous steps are not used for training in subsequent steps and all classes belong to the same domain. Not only does MDIL have a domain drift between any two consecutive steps, there are no restrictions on the label spaces which may or may not have common labels (Table \ref{Tab: problem setting}). 

\vspace{-3pt}
\subsection{Domain Adaptation} 
\vspace{-5pt}
Our work may also be related in a sense to domain adaptation~\cite{zhao2019multi,vu2019advent,yang2020fda,mei2020instance, wu2019ace} 
for semantic segmentation. \cite{wulfmeier2018incremental} proposed incremental unsupervised domain adaptation and showed the effectiveness of learning domain shift by adapting the model incrementally over smaller, progressive domain shifts. Class-incremental domain adaptation~\cite{kundu2020class} focused on source-free domain adaptation while also learning novel classes in target domain. However, all such efforts tackle domain adaptation where source knowledge is adapted to target domains in general, unlike our work in IL which focuses on retaining source domain performance while learning on the target domain (Table \ref{Tab: problem setting}). 



\begin{figure*}[t]
\begin{center}
\includegraphics[width=\textwidth]{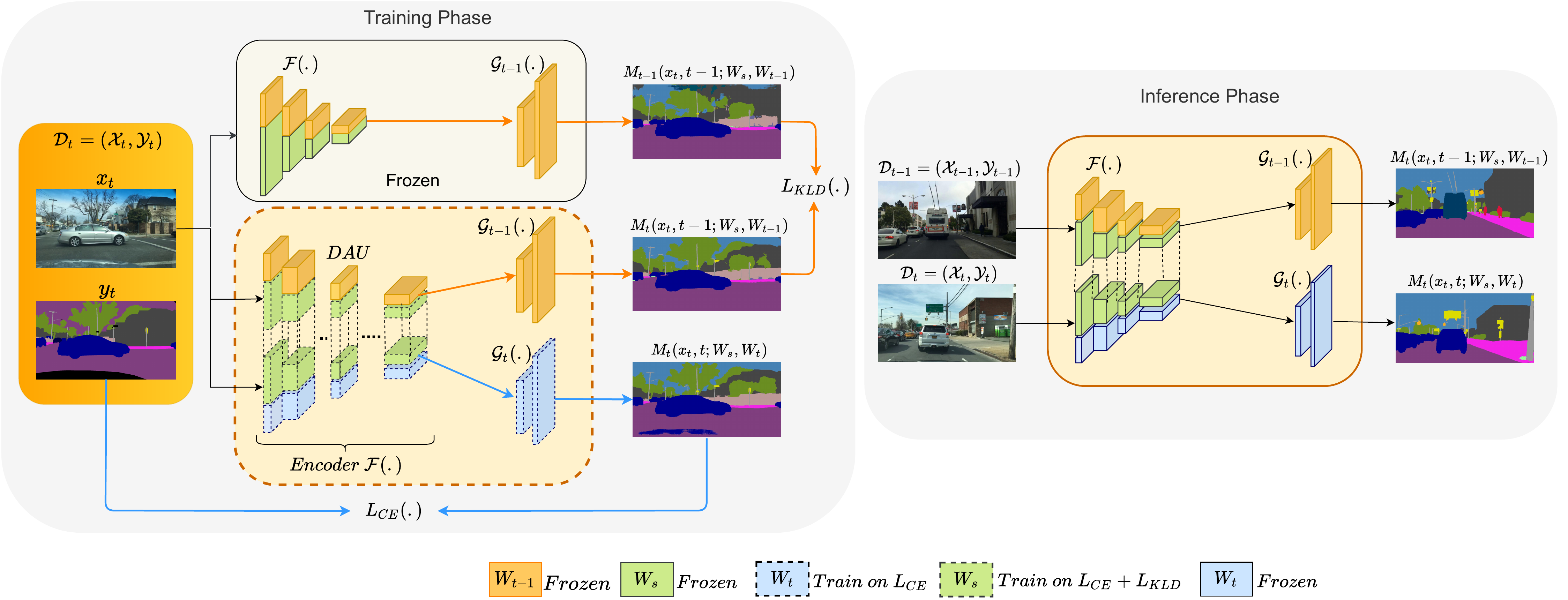}
\end{center}
\vspace{-4mm}
\caption{\textbf{Multi-domain incremental semantic segmentation framework.} \textit{Training Phase:} Training on domain $D_t$ in incremental step $t$. Our model consists of domain-specific decoders and a single encoder. The encoder is composed of Domain-Aware Residual Units (DAU), illustrated in Figure~\ref{fig:DAU}. Layers indicated in \textcolor{YellowGreen}{green} inside the encoder ($W_s$) are common to all domains. They have been shown separately for illustration of separate domain-specific paths. Domain-specific layers of current domain ($W_t$) are in \textcolor{Cerulean}{blue}; domain-specific layers of previous domain ($W_{t-1}$) are in \textcolor{orange}{orange}. \textit{Inference phase:} For evaluation on a particular domain, the corresponding domain-specific path is used to get the segmentation output.}
\vspace{-10pt}
\label{fig:diagram}
\end{figure*}


\begin{figure}[t]
\includegraphics[width=\linewidth]{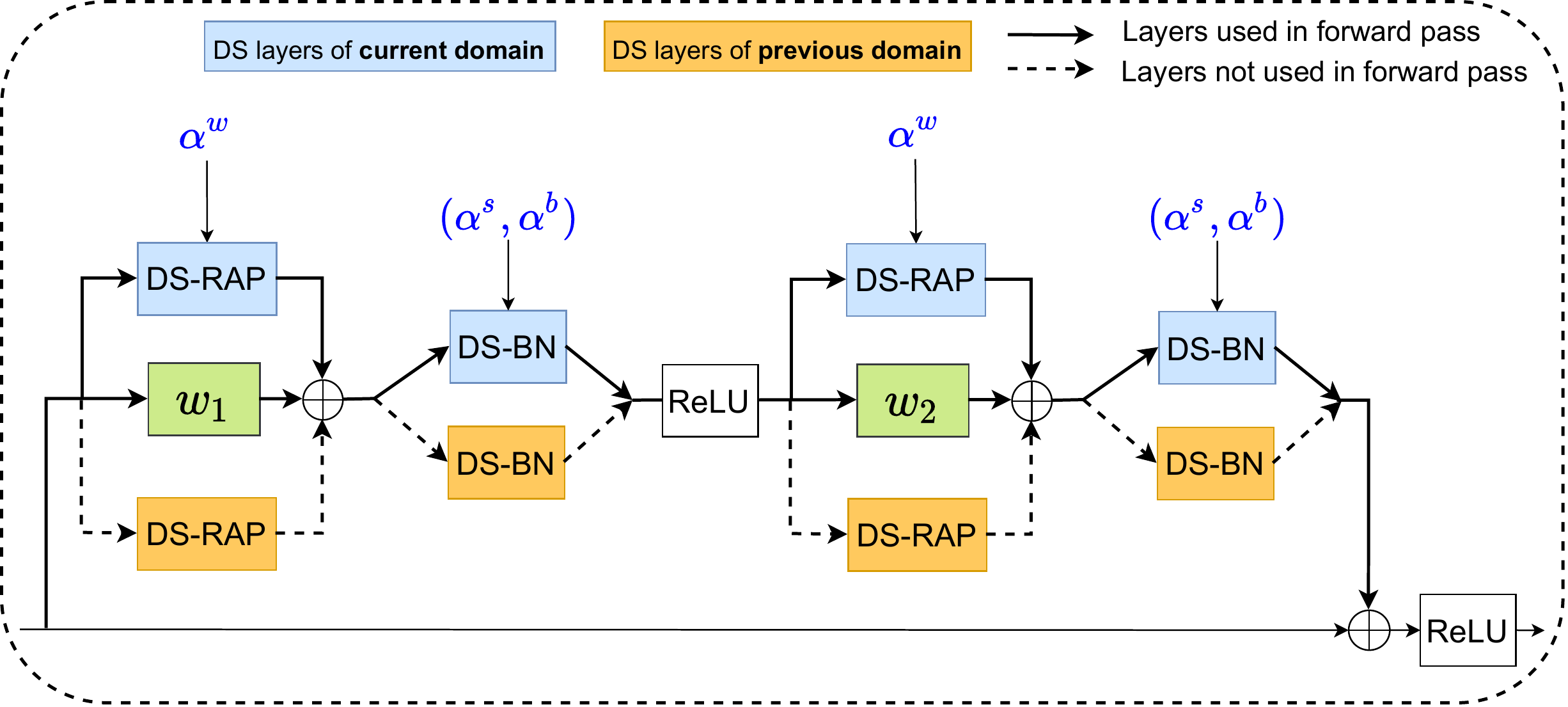} 
\vspace{-4mm}
\caption{\textbf{Domain-Aware Residual Unit (DAU).} These units constitute encoder $\mathcal{F}(\cdot)$. For testing current domain, domain-specific layers for current domain (\textcolor{Cerulean}{blue}) and shared layers (\textcolor{YellowGreen}{green}) are used for forward pass through DAU.}  
\vspace{-12pt}
\label{fig:DAU}
\end{figure}

\vspace{-3pt}
\section{Multi-Domain Incremental Learning for Semantic Segmentation: Methodology}
\vspace{-3pt}

\noindent \textbf{Problem Setting.}
In incremental learning, $\mathcal{T}$ tasks are presented sequentially, each corresponding to a different dataset of domains $\mathcal{D}_1, \mathcal{D}_2,...,\mathcal{D}_t,...,\mathcal{D}_T$ having label spaces $\mathcal{Y}_1, \mathcal{Y}_2,...,\mathcal{Y}_t,..., \mathcal{Y}_T$, respectively. Learning takes place in \textit{incremental steps}, where each step involves learning an existing model on the current task $\mathcal{T}_t$, which in our case is $\mathcal{D}_t$. A domain $\mathcal{D}_t$ represents image data collected from a particular geographic road environment, and $\mathcal{Y}_t$ represents the semantic label space of the classes present in that domain. We consider the general case of non-overlapping label spaces such that the label space $\mathcal{Y}_t$ will either have a full overlap or a partial overlap w.r.t. $\mathcal{Y}_{t-1}$. $\mathcal{Y}_t$ may contain novel classes that were not present in $\mathcal{Y}_{t-1}$ and $\mathcal{Y}_{t-1}$ may also have novel classes absent in $\mathcal{Y}_t$. $\mathcal{D}_t$ has a domain shift with respect to $\mathcal{D}_{t-1}$, as expected.   

Our goal is to train a single semantic segmentation model $M$ that learns to classify data on each domain $\mathcal{D}_t$, sequentially. Hence, given $\mathcal{T}$ tasks, at each IL step $t$, we aim to learn a task-aware mapping $M_t(\mathcal{X}_t, t) = \mathcal{Y}_t$ for the $t^{th}$ domain $\mathcal{D}_t = (\mathcal{X}_t, \mathcal{Y}_t)$, such that performance on any of the previous domains $\mathcal{D}_{t-i}, \ 0<i<t$ does not degrade when learning on the $t^{th}$ domain. 
At any given step $t$, input $\mathcal{X}_{t-i}$ or annotation $\mathcal{Y}_{t-i}$ data pertaining to any previous domain $\mathcal{D}_{t-i}$ is not available for training. Note that we use the terms \textit{task}, \textit{domain}, and \textit{dataset} interchangeably to refer to $\mathcal{D}_t$.

\vspace{3pt}
\noindent \textbf{Proposed Framework.}
Our framework $M$, as illustrated in Figure~\ref{fig:diagram}, is composed of a shared encoder module $\mathcal{F}$ and different domain-specific decoder modules $\mathcal{G}_t$ for prediction in the domain-specific label spaces.
For a given input image $x_t \in \mathcal{D}_t$ at incremental step $t$, our method learns a mapping $M_t(x_t, t; \mathcal{W}_s, \mathcal{W}_t) = \mathcal{G}_t(\mathcal{F}(x_t, t; \mathcal{W}_s, \alpha_t))$, composed of a set of shared, domain-invariant parameters $\mathcal{W}_s$ which are universal for all domains and a domain-specific set of parameters $\mathcal{W}_t$ which are exclusive to its respective domain $\mathcal{D}_t$. The idea is to factorize the network latent space such that homogeneous semantic representation among all datasets gets captured in the shared parameters $\mathcal{W}_s$. In contrast, heterogeneous dataset statistics are learned by the corresponding domain-specific layers $\mathcal{W}_t$. This way, \textit{by construction}, we get a good stability-plasticity trade-off.
Our approach is designed for segmentation models with a ResNet~\cite{he2016deep} based encoder backbone. We modify each residual unit in the encoder to a Domain-Aware Residual Unit (DAU). 

\vspace{3pt}
\noindent \textit{\uline{Domain-Aware Residual Unit:}}
As shown in Figure~\ref{fig:DAU}, each DAU consists of (i) a set of domain-invariant parameters $W_s = \{w_1, w_2\}$, and (ii) a set of domain-specific parameters for each task $t$ given as, $\alpha_t = \{\alpha^w, \alpha^s, \alpha^b\}$. ${w_1, w_2}$ are the $3 \times 3$ convolutional layers present in a traditional residual unit~\cite{he2016deep} and are shared among all domains. Domain-specific layers in the DAU are of two kinds: (i) Domain-Specific Parallel Residual Adapter layers (DS-RAP), and (ii) Domain-Specific Batch Normalization layers (DS-BN). We use the concept of Parallel Residual Adapters (RAP) from~\cite{rebuffi2018efficient} and modify its optimization for our setting. DS-RAP layers, $\alpha^w$ are $1\times1$ convolutional layers added to the shared convolutional layers in parallel. They act as layer-level domain adapters. Differently from the residual adapter module in~\cite{rebuffi2018efficient}, we make the Batch Normalization layers also domain-specific. In BN, the normalized input is scaled and shifted as $s \odot x +  b$; here, $(\alpha^s, \alpha^b)$ denote the learnable scale and shift parameters of the DS-BN layers. The shared weights act like universal filter banks, learning domain-generalized knowledge. In contrast, the DS-RAP and DS-BN layers are exclusive to their particular domain, responsible for learning domain-specific features. 


Existing residual adapter-based approaches for image classification~\cite{kanakis2020reparameterizing,rebuffi2017learning, rebuffi2018efficient} freeze their shared parameters $W_s$ to a generic initialization such as Imagenet~\cite{deng2009imagenet} pre-trained weights and train only domain-specific parameters. We find that Imagenet initialization for $W_s$ does not work well for fine-grained tasks like semantic segmentation (Section \ref{ablations}). Thus, instead of freezing shared parameters $W_s$, we fine-tune them on the new domain $\mathcal{D}_t$, in an end-to-end training. We propose an optimization strategy that makes $W_s$ parameters learn domain-agnostic features and a different optimization for $W_t$ parameters to make them domain-specific, as described below.

\vspace{3pt}
\noindent \textbf{Optimization Strategy.}
\vspace{3pt}
\noindent \textit{\uline{Domain-specific parameters:}}
For a particular task $t$, the composition of domain-specific parameters is given as $W_t = \{\alpha_t, \mathcal{G}_t\}$. To learn a new domain $\mathcal{D}_t$ at step $t$, we add new parameters $W_t$ to the model from the previous step $M_{t-1}$ and call this model $M_{t}$. We initialize all $W_t$ from $W_{t-1}$, except the output classifier layer which is randomly initialized (label space $\mathcal{Y}_t$ may be different from $\mathcal{Y}_{t-1}$). We refer to this initialization strategy as \textit{$init_{W_t}$}. The domain-specific layers $W_t$ are trained only on the task-specific loss for domain $\mathcal{D}_t$ given as: 
\vspace{-4pt}
\begin{equation} \label{eq:1}
    L_{CE_t} = \frac{1}{N} \sum_{x_t \epsilon \mathcal{D}_t} \psi_t(y_t, \mathcal{G}_t(\mathcal{F}(x_t, t; \mathcal{W}_s, \alpha_t)) 
\vspace{-4pt}
\end{equation}
\noindent where $\psi_t$ is the task-specific softmax cross-entropy loss function over the label space $\mathcal{Y}_t$. All domain-specific layers of previous domains $W_{t-i}, 0 < i <t$ remain frozen during current domain training. 

\vspace{3pt}
\noindent \textit{\uline{Domain-invariant parameters:}}
The $W_s$ layers in the encoder are shared among all tasks. In IL step $t$, we initialize these weights from the corresponding weights in $M_{t-1}$. In addition to the task-specific cross entropy loss $L_{CE_t}$, we use a regularization loss $L_{KLD}$ to optimize the shared weights:
\vspace{-4pt}
\begin{equation} \label{eq:2}
    q_i^s = M_t(x_t, t-1; W_s, W_{t-1}) 
\vspace{-4pt}
\end{equation}
\begin{equation} \label{eq:3}
     q_i^t = M_{t-1}(x_t, t-1; W_s, W_{t-1})
\vspace{-4pt}
\end{equation}
\begin{equation} \label{eq:4}
     L_{KLD} = \lambda_{KLD} \cdot \sum_{i=1}^{t-1} \sum_{x_t \epsilon \mathcal{D}_t} \phi(q_i^s, q_i^t) 
\vspace{-4pt}
\end{equation}
\noindent where $q_i^s$ is the prediction map of the current model $M_t$ on the current input $x_t$ for the \textit{previous task $t-1$}; $q_i^t$ is the prediction map of the previous model $M_{t-1}$ on the current input $x_t$ for the \textit{previous task $t-1$}; $\phi$ is the KL-divergence (KLD) loss between these two softmax probability distribution maps, computed and summed over each previously learned task $i$, $0 < i < t$; $\lambda_{KLD}$ is the regularization hyperparameter for KL-divergence. KLD here effectively \textit{distills domain knowledge} from the teacher $q_i^t$ to the student $q_i^s$, and can be seen as domain adaptive knowledge distillation~\cite{kothandaraman2021domain}. 
Total loss for domain-invariant parameters $W_s$ is hence given as:
\vspace{-4pt}
\begin{equation} \label{eq:5}
    L_{W_{s}} = L_{CE_t} + L_{KLD}
\vspace{-4pt}
\end{equation}
Optimization of the $W_s$ and $W_t$ parameters in our model is at a differential learning rate, \textit{dlr}. We observe that optimizing $W_s$ at a learning rate $100\times$ lower than the learning rate of $W_t$ stabilizes $W_s$ w.r.t. $W_{t-1}$ and prevents forgetting.
As shown in Section~\ref{ablations}, a combination of the \textit{$init_{W_t}$} and \textit{dlr} learning strategies is responsible for preserving old knowledge and learning new knowledge simultaneously. We summarize the optimization protocol in Algorithm \ref{alg:cap}.

When the shared weights $W_s$ are trained on the domain-specific loss $L_{CE_t}$ of the current step, they learn the current domain's features and quickly forget the domain-specific representation learned on the previous domain. The \textit{dlr} strategy prevents the shared weights $W_s$ from drifting away from the domain-specific features learned in the previous step, when learning the current domain. This model is referred to as DAU-FT-\textit{dlr} in our results (Table~\ref{Tab: ablation studies}). Minimizing KLD between the output feature maps of the previous and current models 
\textit{preserves previous tasks' domain knowledge in the shared weights}. A combination of $L_{KLD}$, $L_{CE_t}$ and \textit{dlr} learning strategy thus help train domain-invariant shared layers in the encoder. $W_t$ weights are domain-specific as they are trained only on the domain-specific loss. Together, the above steps reparametrize the model into domain-specific and domain-invariant features, which in turn achieve strong performance on the new domain while retaining performance on older domains. 

\vspace{3pt}
\noindent \textbf{Inference Phase.}
For a query image $x_t \in \mathcal{D}_t, t \in T$ (set of tasks the model has learned on so far), our model gives an output segmentation map of pixel-wise predictions $\hat{y}_t = M_t(x_t, t)$ over the label space $\mathcal{Y}_t$. For evaluation on any domain $\mathcal{D}_t$, only the corresponding domain-specific $\alpha_t$ and $\mathcal{G}_t$ get activated in the forward pass. 
In effect, our model has multiple domain-specific paths with a large degree of parameter sharing. 

\vspace{3pt}
\noindent \textbf{Experimental Setting.}
Consider a two-task IL setting where the goal is to take a model trained on geographical domain $D_A$ and incrementally learn on another domain $D_B$. Multi-domain incremental semantic segmentation consists of two scenarios: (i) Case 1: $D_A$ and $D_B$ have a domain shift, but aligned label spaces, i.e. $D_A \neq D_B, \mathcal{Y}_A = \mathcal{Y}_B$; (ii) Case 2: $D_A$ and $D_B$ have a domain shift as well as non-overlapping label spaces, i.e. $D_A \neq D_B, \mathcal{Y}_A \neq \mathcal{Y}_B$. Our proposed approach for MDIL tackles both these scenarios.

\algnewcommand\algorithmicinput{\textbf{Initialize:}}
\algnewcommand\Initialize{\item[\algorithmicinput]}

\algnewcommand\algorithmicupdate{\textbf{Update:}}
\algnewcommand\Update{\item[\algorithmicupdate]}

\algnewcommand\algorithmicfreeze{\textbf{Freeze:}}
\algnewcommand\Freeze{\item[\algorithmicfreeze]}
\begin{algorithm}[t]
\footnotesize
\caption{Training protocol in the $t^{th}$ incremental step}\label{alg:cap}
\begin{algorithmic}
\Require
\State \hskip0em $\mathcal{D}_t$: new data of current step $t$ 
\State \hskip0em $M_{t-1}$: model from previous step $t-1$

\Initialize
\State \hskip0em $M_t \gets$ add new DS layers $W_t$ to $M_{t-1}$ (for learning $\mathcal{D}_t$)
\State \hskip0em \textit{$init_{W_t}$}: $W_t$ in $M_t \gets W_{t-1}$ in $M_{t-1}$
\Freeze{DS weights of all previous domains; $W_{t-i}, 0 < i <t$}
\end{algorithmic}
\begin{algorithmic}[1]
\For{epochs}
\For{mini-batch}
\State Forward pass $M_t(x_t, t)$ via $W_t$ 
\State Compute task-specific loss $L_{CE_t}$ for $\mathcal{D}_t$ by Eq. \ref{eq:1}
\State Forward pass $M_t(x_t, t-1)$ via $W_{t-1}$, Eq. \ref{eq:2}
\State Forward pass $M_{t-1}(x_t, t-1)$ via $W_{t-1}$ of $M_{t-1}$, Eq. \ref{eq:3}
\State Compute KLD loss $L_{KLD}$ by Eq. \ref{eq:4}
\State Compute $L_{W_{s}}$ by Eq. \ref{eq:5}
\State {\bf Update:}
\State \hskip1em $L_{CE_t}$ on $W_t$ at standard network learning rate \textit{lr}
\State \hskip1em $L_{W_{s}}$ on $W_s$ at a lower learning rate \textit{dlr}
\EndFor
\EndFor
\end{algorithmic}
\begin{algorithmic}
\State \hskip-1em Discard training data $\mathcal{D}_t$
\end{algorithmic}
\end{algorithm}
\vspace{-5pt}

\begin {table*}[ht]
\centering
\small
    \begin {tabular} {p{2.4cm}|c|c|c|c|c|c|c|c}
    \hline
    IL Step & \multicolumn{2}{c|}{Step 1} & \multicolumn{3}{c|}{Step 2: $D_A \neq D_B, \mathcal{Y}_A = \mathcal{Y}_B$} & \multicolumn{3}{c}{Step 2: $D_A \neq D_B, \mathcal{Y}_A \neq \mathcal{Y}_B$} \\
     & \multicolumn{2}{c|}{$CS$} & \multicolumn{3}{c|}{$CS \rightarrow BDD$} & \multicolumn{3}{c}{$CS \rightarrow IDD$} \\
    \hline
    Methods & CS $\uparrow$ & $\Delta_m\% \downarrow$ & CS $\uparrow$ & BDD $\uparrow$ & $\Delta_m\% \downarrow$ & CS $\uparrow$ & IDD $\uparrow$ & $\Delta_m\% \downarrow$ \\
    \hline
    Single-task & 72.55 & & 72.55 & 54.1 & & 72.55 & 61.97 & \\
    Multi-task & 72.55 & & 69.42 & 57.69 & 1.16\% $(\uparrow)$ & 71.11 & 60.85 & 1.89\% \\
    \hline
    FT & 72.55 & 0.0\% & 40.05 \textcolor{red}{(-32.5)} & 52.74 & 23.66\% & 36.81 \textcolor{red}{(-35.74)} & 61.56 & 24.96\% \\
    FE & 72.55 & 0.0\% & 72.55 \textcolor{red}{(-0.00)} & 42.93 & 10.32\% & 72.55 \textcolor{red}{(-0.00)} & 45.69 & 13.14\% \\
    \hline
    FT (Single-Head) & 72.55 &  & 47.42 \textcolor{red}{(-25.13)} & 50.89 & 20.29\% & 36.82 \textcolor{red}{(-35.73)} & 53.79 & 31.22\% \\ 
    LwF~\cite{li2017learning} & 72.55 &  & 58.66 \textcolor{red}{(-13.89)} & 43.26 & 19.59\% & 62.63 \textcolor{red}{(-9.92)} & 42.89 & 22.23\%  \\ 
    ILT~\cite{michieli2019incremental} & 72.55 &  & 56.84 \textcolor{red}{(-15.71)} & 32.97 & 30.36\% & 54.37 \textcolor{red}{(-18.18)} & 25.07 & 42.30\%  \\ 
    \hline
    \rowcolor{Gainsboro!60}
    Ours & 71.82 & 1.01\% & 65.21 \textcolor{red}{(-7.34)} & 55.73 \textcolor{blue}{(+1.63)} & \textbf{3.55\%} & 64.58 \textcolor{red}{(-7.97)} & 59.11 \textcolor{red}{(-2.86)} & \textbf{7.80\%} \\
    \hline
\end{tabular}
\caption{\textit{Results of 2-task incremental settings.}
We report performance on all datasets, after incrementally learning on the current dataset $D_t$ in step $t$. Arrows indicate order of learning. Parenthesis show drop/gain in performance w.r.t single-task baseline for the corresponding dataset. Lower $\Delta_m\%$ indicates better stability-plasticity trade-off and overall performance.}
\vspace{-11pt}
\label{Tab: main result table step2}
\end{table*}

\begin {table}[t]
\centering
\footnotesize	
\begin{adjustbox}{width=0.47\textwidth}
    \begin {tabular} {p{1.3cm}|c|c|c|c}
    \hline
    IL Step & \multicolumn{4}{c}{Step 3: $D_A \neq D_B, \mathcal{Y}_A \neq \mathcal{Y}_B$} \\
     & \multicolumn{4}{c}{$CS \rightarrow BDD \rightarrow IDD$} \\
    \hline
    Methods & CS $\uparrow$ & BDD $\uparrow$ & IDD $\uparrow$ & $\Delta_m\%$ $\downarrow$ \\
    \hline
    Single-task & 72.55 & 54.1 & 61.97 & \\
    Multi-task & 69.37 & 58.13 & 59.37 & 0.38\% \\
    \hline
    FT & 30.49 \textcolor{red}{(-42.06)} & 32.05 \textcolor{red}{(-22.05)} & 60.65 & 33.62\% \\
    FE & 72.55 \textcolor{red}{(-0.00)} & 42.93 \textcolor{red}{(-11.17)} & 46.09 & 15.42\% \\
    \hline
    \rowcolor{Gainsboro!60}
    Ours & 59.19 \textcolor{red}{(-13.36)} & 49.66 \textcolor{red}{(-4.44)} & 59.16 & \textbf{10.39\%} \\
    \hline
    \hline
     & \multicolumn{4}{c}{$CS \rightarrow IDD \rightarrow BDD$} \\
    \hline
    Methods & CS $\uparrow$ & IDD $\uparrow$ & BDD $\uparrow$ & $\Delta_m\%$ $\downarrow$ \\
    \hline 
    Single-task & 72.55 & 61.97 & 54.1 & \\
    Multi-task & 69.37 & 59.37 & 58.13 & 0.38\% \\
    \hline
    FT & 36.19 \textcolor{red}{(-36.36)} & 26.3 \textcolor{red}{(-35.67)} & 53.37 & 36.34\% \\
    FE & 72.55 \textcolor{red}{(-0.00)} & 45.69 \textcolor{red}{(-16.28)} & 43.06 & 15.56\% \\
    \hline
    \rowcolor{Gainsboro!60}
    Ours & 62.55 \textcolor{red}{(-10.0)} & 53.85 \textcolor{red}{(-8.12)} & 55.90 & \textbf{7.85\%} \\
    \hline
\end{tabular}
\end{adjustbox}
\caption{\textit{Results of 3-task incremental learning settings.} $CS \rightarrow BDD \rightarrow IDD$ model was trained on CS in step 1, on BDD in step 2. Performance is reported on all 3 datasets after it is incrementally trained on IDD in step 3. Similarly, we report results on the $CS \rightarrow IDD \rightarrow BDD$ setting.}
\vspace{-10pt}
\label{Tab: main result table step3}
\end{table}
\vspace{-3pt}
\section{Experiments and Results} \label{comparisons on 2 scenarios}
\vspace{-3pt}

\noindent \textbf{Datasets.}
We perform IL over three highly diverse, large-scale urban driving datasets collected from different geographic locations. The Cityscapes dataset (CS)~\cite{cordts2016cityscapes} is a standard autonomous driving dataset of daytime images collected from urban streets of 50 European cities. It contains 19 labels, captured in 2975 training and 500 validation images. The Berkeley Deep Drive dataset (BDD) is a widespread collection of road scenes spanning diverse weather conditions and times of the day in the United States~\cite{yu2020bdd100k}. It covers residential areas and highways along with urban streets. It has 7000 training and 1000 validation images. The Indian Driving Dataset (IDD) has unconstrained road environments collected from Indian cities~\cite{varma2019idd}. These road environments captured in 6993 training and 781 validation images are highly unstructured with unique labels like billboard, auto-rickshaw, animal, etc. In our experiments, we adhere to the default label spaces of these datasets, i.e., we use the 19 labels in Cityscapes and BDD100k, and IDD level 3, which has 26 labels.

\vspace{3pt}
\noindent \textbf{Evaluation Metrics.}
We use the mean Intersection-over-Union (mIoU) metric to evaluate the semantic segmentation performance of a model on each dataset, following standard practice. Similar to~\cite{kanakis2020reparameterizing}, we quantify the overall IL performance of a model $m$, as the average per-task drop in semantic segmentation performance (mIoU) with respect to the corresponding single-task baseline $b$:
\vspace{-4pt}
\begin{equation}
    \Delta_m\% = \frac{1}{T} \sum_{t=1}^{T} \frac{mIoU_{m,t} - mIoU_{b,t}}{mIoU_{b,t}} 
\vspace{-4pt}
\end{equation}
where $mIoU_{m,t}$ is the segmentation accuracy of model $m$ on task $t$. $\Delta_m\%$ quantifies the stability-plasticity trade-off to give an overall score of IL performance. 

\begin {table}[t]
\footnotesize	
\begin{adjustbox}{width=0.47\textwidth}
    \begin {tabular}{p{1.3cm}|c|c|c|c|c}
    \hline
    IL Step & \multicolumn{2}{c|}{Step 1} & \multicolumn{3}{c}{Step 2: $D_A \neq D_B, \mathcal{Y}_A \neq \mathcal{Y}_B$} \\ 
     & \multicolumn{2}{c|}{$IDD$} & \multicolumn{3}{c}{$IDD \rightarrow BDD$} \\
    \hline
    Methods & IDD $\uparrow$ & $\Delta_m\% \downarrow$ & IDD $\uparrow$ & BDD $\uparrow$ & $\Delta_m\% \downarrow$ \\
    \hline
    Single-task 							
    & 61.97 & & 61.97 & 54.1 & \\
    Multi-task & 61.97 &  & 61.05 & 56.05 & 1.06\% ($\uparrow$) \\
    \hline 
    FT & 61.97 & 0.0 & 27.33 & 52.88 & 29.08\% \\ 
    FE & 61.97 & 0.0 & 61.97 & 46.23 & 7.27\% \\
    \hline
    \rowcolor{Gainsboro!60}
    Ours & 62.60 & 1.02 ($\uparrow$) & 57.36 & 55.73 & \textbf{2.21\%} \\
    \hline
    \hline
     & \multicolumn{2}{c|}{$BDD$} & \multicolumn{3}{c}{$BDD \rightarrow IDD$} \\
    \hline
    Methods & BDD $\uparrow$ & $\Delta_m\% \downarrow$ & BDD $\uparrow$ & IDD $\uparrow$ & $\Delta_m\% \downarrow$ \\
    \hline
    Single-task & 54.1 &  & 54.1 & 61.97 & \\
    Multi-task & 54.1 &  & 56.05 & 61.05 & 1.06\% ($\uparrow$) \\
    \hline
    FT & 54.1 & 0.0 & 30.72 & 59.9 & 23.28\% \\
    FE & 54.1 & 0.0 & 54.1 & 47.24 & 11.88\% \\
    \hline
    \rowcolor{Gainsboro!60}
    Ours & 52.1 & 3.70 & 50.92 & 57.21 & \textbf{6.78\%} \\
    \hline
\end{tabular}
\end{adjustbox}
\caption{Results of domain ordering on $IDD \rightarrow BDD$ and $BDD \rightarrow IDD$ 2-task incremental settings.}
\vspace{-14pt}
\label{Tab: domain ordering}
\end{table}

\vspace{3pt}
\noindent \textbf{Implementation Details.}
We use ERFNet~\cite{romera2017erfnet} as the backbone for implementing this work, as it allows the dynamic addition of our modules seamlessly.
Similar to~\cite{kalluri2019universal, wang2021cross, cermelli2020modeling}, we report mIoU on the standard validation sets of these datasets. More details are provided in the supplementary.

\vspace{-1pt}
\begin{figure*}[t]
\begin{center}
\includegraphics[width=0.99\textwidth]{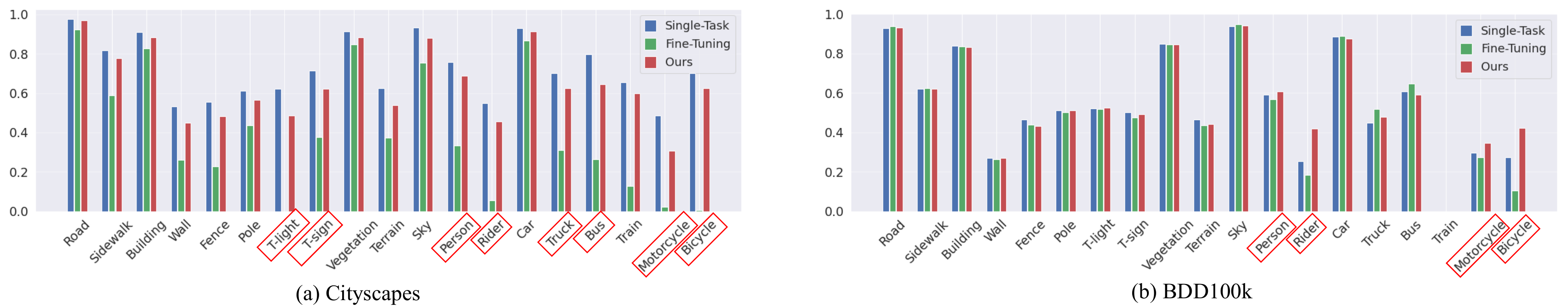} 
\end{center}
\vspace{-14pt}
\caption{Analysis of class-wise accuracy (IoU) on (a) CS and (b) BDD after incrementally learning from CS $\rightarrow$ BDD. Our model is able to show significant improvement for the categories marked in red.}
\vspace{-11pt}
\label{fig:class-wise-iou}
\end{figure*}

\vspace{3pt}
\noindent \textbf{Results.}
We show detailed analysis on two, 2-task settings $CS \rightarrow BDD$ and $CS \rightarrow IDD$ to compare the two possible cases. We also show results on 3-task settings including, $CS \rightarrow BDD \rightarrow IDD$ and $CS \rightarrow IDD \rightarrow BDD$.  

\vspace{3pt}
\noindent \textit{\uline{Incremental Learning Baselines:}}
We compare our proposed approach with four standard IL baselines. The single-task baseline denotes datasets trained independently on separate models, which one can consider as the gold standard or the upper bound for IL performance. This is used to compute catastrophic forgetting and overall evaluation score $\Delta_m\%$ across our experiments. The multi-task model gives the joint training performance, where a single multi-decoder model is trained offline on all the datasets together (note that this violates the IL setting, but is shown for completeness). Fine-tuning (FT) is a standard baseline in IL, where a model is fine-tuned on the newer domain without any explicit effort to mitigate forgetting. 
This can be considered as a lower bound for our experiments. In feature extraction (FE), we freeze all encoder weights and only train the new domain's decoder weights. Fine-tuning gives the maximum plasticity and minimum stability, while feature extraction exhibits maximum stability and minimum plasticity. We also compare our method against an existing class-IL method for segmentation, ILT~\cite{michieli2019incremental}. This is a single-head architecture comparison along with the fine-tuning (single-head). We also compare against learning without forgetting LwF~\cite{li2017learning} implemented as a multi-head. 


\noindent \textit{ \uline{Cityscapes $\rightarrow$ BDD100k :}}
In this setting, we start by learning a model on Cityscapes (CS) in step 1, followed by incrementally learning the same model on BDD100k (BDD) in step 2. CS and BDD datasets have a common label space of 19 labels. Hence, there is a domain shift when going from CS $\rightarrow$ BDD, but their label spaces are aligned. As shown in Table~\ref{Tab: main result table step2}, using our model, forgetting on CS has been mitigated by 25.16\% (w.r.t the fine-tuning baseline) and is only 7.34\% below the single-task upper limit. A comparison of the class-wise performance of our approach with the fine-tuning baseline is given in Figure~\ref{fig:class-wise-iou}. 
Our model mitigates forgetting in all 19 classes, and retains performance by a significantly large margin ($\geq$ 30\%) on \textit{safety-critical classes} such as traffic light, traffic sign, person, rider, truck, bus and bicycle. 
Importantly, we observe that our proposed model has surpassed the single-task model performance on BDD by 1.63\%. We hypothesize that this forward transfer is achieved since our model captures the domain-specific characteristics of the dataset distributions of CS and BDD in the domain-specific parameters. 
Class-wise analysis are explained in detail in supplementary material. 

\vspace{3pt}
\noindent \textit{\uline{Cityscapes $\rightarrow$ IDD :}}
As presented in Table~\ref{Tab: main result table step2}, we first learn a model on CS in step 1, then incrementally learn on IDD in step 2. CS has a label space of 19 labels, and IDD has a label space of 26 labels such that a subset of 17 road classes is common. 2 classes are exclusive to CS, while 8 classes are exclusive to IDD. This scenario includes both a domain shift as well as label misalignment. Forgetting on CS is mitigated by 27.77\% by our model. This shows that despite the label misalignment, our approach can retain old task performance in CS $\rightarrow$ IDD almost as well as it does in the CS $\rightarrow$ BDD setting (forgetting on CS is 7.97\% after learning on IDD as compared to the 7.34\% after learning on BDD). 

\vspace{3pt}
\noindent \textit{\uline{3-Task Incremental Settings :}}
In Table \ref{Tab: main result table step3}, we show results for Cityscapes $\rightarrow$ BDD100k $\rightarrow$ IDD and Cityscapes $\rightarrow$ IDD $\rightarrow$ BDD100k settings.
We also explore different sequences of domain ordering in Table \ref{Tab: domain ordering}. These results show that our model is generalizable with respect to domain ordering. More exhaustive permutations are given in supplementary.

\begin{table}[t]
\begin{center}
\small
\resizebox{0.478\textwidth}{!}{%
\begin {tabular}{p{0.1\textwidth}|c|c|c|c|c|c}
    \hline
    Methods & $L_{KLD}$ & \textit{dlr} & \textit{$init_{W_t}$} & DAU & CS & BDD \\
    \hline
    Single-task & $\times$ & $\times$ & $\times$ & $\times$ & 72.55 & 54.1 \\
    \hline
    DAU-FT & $\times$ & $\times$ & $\times$ & \checkmark & 1.34 & 49.96 \\
    DAU-FT-\textit{dlr}1 & $\times$ & $\times$ & \checkmark & \checkmark & 8.41 & 54.51 \\ 
    DAU-FT-rinit & $\times$ & \checkmark & $\times$ & \checkmark & 46.4 & 54.50 \\
    DAU-FT-\textit{dlr} & $\times$ & \checkmark & \checkmark & \checkmark & 58.4 & 57.03 \\
    \hline
    Ours & \checkmark & \checkmark & \checkmark & \checkmark & 65.21 & 55.73 \\
    \hline
    \end{tabular}%
    }
    \end{center}
    \vspace{-8pt}
    \caption{Ablation studies on the contribution of each component of our proposed model for the $CS \rightarrow BDD$ setting. mIoU on CS, BDD is reported after learning on BDD.}
    \vspace{-11pt}
    \label{Tab: ablation studies}
\end{table}

\vspace{-4pt}
\section{Ablation Studies and Analysis} \label{analysis} 
\label{ablations}
\vspace{-3pt}
\noindent \textbf{Significance of optimization strategies.} In this section, we study the significance of the optimization and initialization strategy we use for attaining a stability-plasticity trade-off in our IL setting. Table~\ref{Tab: ablation studies} shows these results. In IL step $t$, the $W_s$ and $W_{t-1}$ weights in current model $M_t$ are initialized from the corresponding layers in $M_{t-1}$ (for all experiments). The DS weights of current domain $W_t$ can either be randomly initialized or initialized from the DS weights of previous domain $W_{t-1}$. We call the latter as \textit{$init_{W_t}$}. In DAU-FT model, we perform vanilla fine-tuning of the $W_s$ and randomly initialize $W_t$ weights using the standard learning rate on the task-specific loss $L_{CE_t}$. This performs poorly on both old and new domains, catastrophically forgetting the previous domain (1.34\% mIoU). 

Next, we define a differential learning rate \textit{dlr} as $\frac{LR_{W_t}}{LR_{W_s}}$. If we use \textit{dlr=1} and fine-tune both $W_s$ and $W_t$ using same LR, the $W_s$ parameters learn on the new domain (BDD) and forget the previously learned representation on CS (DAU-FT-\textit{dlr1} in table). All \textit{dlr} models use \textit{$init_{W_t}$} unless stated otherwise. As we decrease the LR of $W_s$ w.r.t $W_t$, stability of the model w.r.t the previous domain increases and plasticity w.r.t. the new domain decreases. We find that a \textit{dlr} value of 100 gives a good stability-plasticity trade-off. This model is referred to as DAU-FT-\textit{dlr} in the table. Applying $L_{KLD}$ on the shared weights $W_s$ of DAU-FT-\textit{dlr} model further mitigates forgetting by 7.91\% and is our proposed model (\textit{ours}). It is important that the DS weights $W_t$ not be randomly initialized. If they are randomly initialized, there is a performance drop as given by DAU-FT-rinit.  


\begin {table*}[htb!]
\begin{center}
\resizebox{\textwidth}{!}{%
    \begin {tabular}{p{2.3cm}|c|c|c|c|c|c|c|c|c|c|c|c}
    \hline
    IL Step & \multicolumn{2}{c|}{Step 1} & \multicolumn{3}{c|}{Step 2: $D_A \neq D_B, \mathcal{Y}_A = \mathcal{Y}_B$} & \multicolumn{3}{c|}{Step 2: $D_A \neq D_B, \mathcal{Y}_A \neq \mathcal{Y}_B$} & \multicolumn{4}{c}{Step 3: $D_A \neq D_B, \mathcal{Y}_A \neq \mathcal{Y}_B$} \\
     & \multicolumn{2}{c|}{$CS$} & \multicolumn{3}{c|}{$CS \rightarrow BDD$} & \multicolumn{3}{c|}{$CS \rightarrow IDD$} & \multicolumn{4}{c}{$CS \rightarrow BDD \rightarrow IDD$} \\
    \hline
    Methods & CS $\uparrow$ & $\Delta_m\% \downarrow$ & CS $\uparrow$ & BDD $\uparrow$ & $\Delta_m\% \downarrow$ & CS $\uparrow$ & IDD $\uparrow$ & $\Delta_m\% \downarrow$ & CS $\uparrow$ & BDD $\uparrow$ & IDD $\uparrow$ & $\Delta_m\%$ $\downarrow$ \\
    \hline
    Single-task & 72.55 & & 72.55 & 54.1 & & 72.55 & 61.97 & & 72.55 & 54.1 & 61.97 & \\
    \hline
    RCM-NFI~\cite{kanakis2020reparameterizing} & 1.98 & 97.27\% & 1.98 & 1.32 & 97.42\% & 1.98 & 1.13 & 97.72\% & 1.98 & 1.32 & 1.13 & 97.67\% \\
    RCM-I~\cite{kanakis2020reparameterizing} & 63.13 & 12.98\% & 63.13 & 47.94 & 12.19\% & 63.13 & 55.66 & 11.58\% & \textbf{63.13} & 47.94 & 55.66 & 11.52\% \\
    RAS-I~\cite{rebuffi2017learning} & 61.65 & 15.02\% & 61.65 & 48.05 & 13.10\% & 61.65 & 54.09 & 13.87\% & 61.65 & 48.05 & 54.09 & 12.97\%  \\
    RAP-I~\cite{rebuffi2018efficient} & 58.43 & 19.46\% & 58.43 & 46.34 & 16.90\% & 58.43 & 50.97 & 18.61\% & 58.43 & 46.34 & 51.27 & 17.02\% \\
    \hline
    \rowcolor{Gainsboro!60}
    Ours & \textbf{71.82} & 1.01\% & \textbf{65.21} & \textbf{55.73} & \textbf{3.55\%} & \textbf{64.58} & \textbf{59.11} & \textbf{7.80\%} & 59.19 & \textbf{49.66} & \textbf{59.16} & \textbf{10.39\%} \\
    \hline
\end{tabular}%
}
\end{center}
\vspace{-11pt}
\caption{Comparison with other residual adapter-based architectures. Lower score $\Delta_m\%$ indicates better overall performance.}
\vspace{-8pt}
\label{Tab: comparison with sota}
\end{table*}


\begin{figure}[ht] 
\begin{center}
\includegraphics[width=0.48\textwidth]{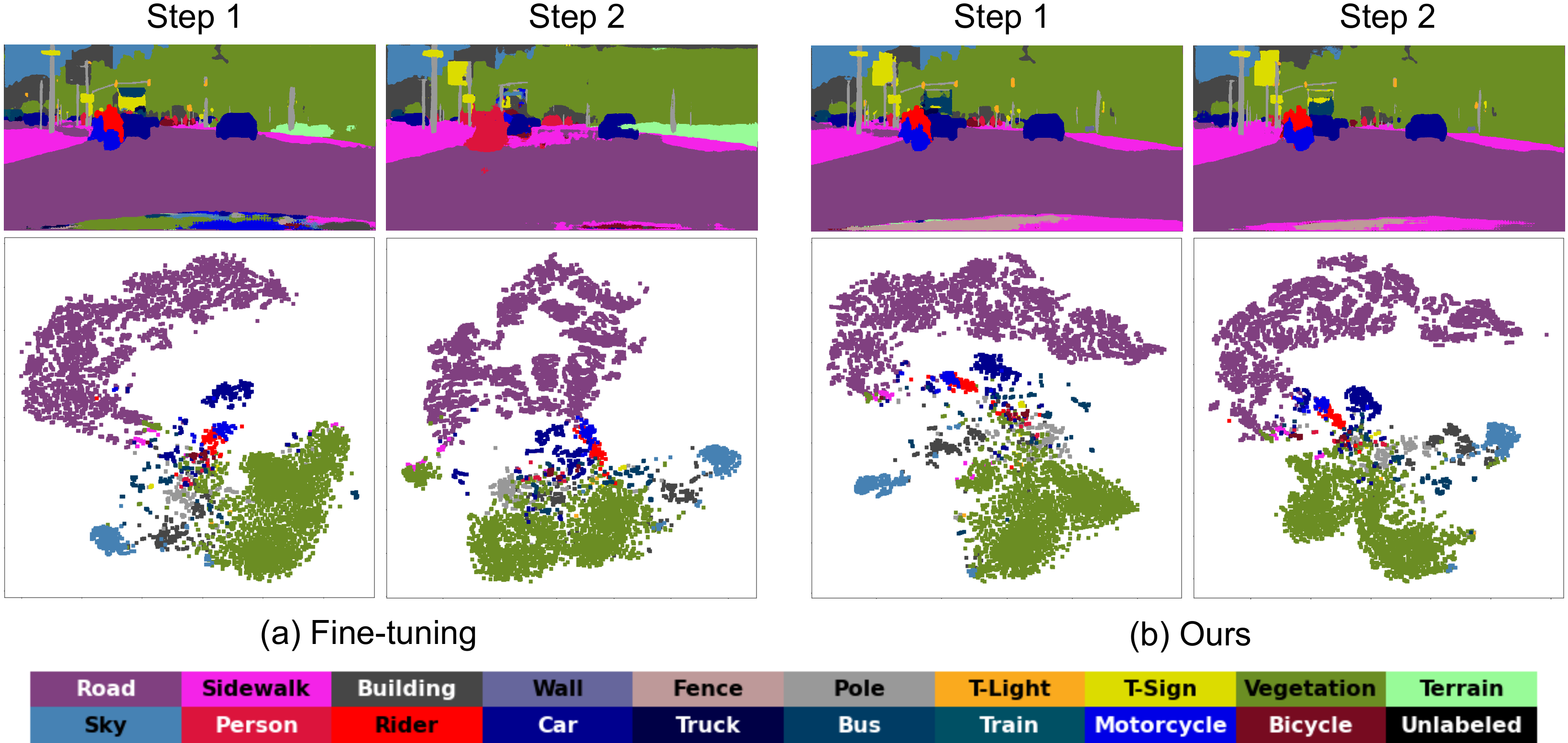} 
\end{center}
\vspace{-6mm}
\caption{Consider the CS $\rightarrow$ BDD setting. (a) t-SNE of CS features for the \textit{fine-tuning} model in step 1 and step 2. (b) t-SNE of CS features for \textit{our} proposed model in step 1 and step 2. Fine-tuning distorts the latent space representation of CS learned in step 1. \textbf{Our model preserves the latent space of CS after learning on BDD in step 2.}}
\vspace{-9pt}
\label{fig:tsne}
\end{figure}

\vspace{3pt}
\noindent \textbf{Comparison with other residual adapter-based architectures.} 
We show a comparison of our approach against three state-of-the-art residual adapter-based
methods in Table~\ref{Tab: comparison with sota}. RAP-I~\cite{rebuffi2018efficient} denotes the parallel residual adapter with $W_s$ weights frozen from pre-training on ImageNet. Our proposed optimization strategy outperforms this by a large margin. RAS-I is the series residual adapter~\cite{rebuffi2017learning}. While RAP and RAS contain layer-level residual adapters $\alpha$ in the shared encoder, RCM (Reparameterized Convolutions for Multi-task learning)~\cite{kanakis2020reparameterizing} is a block-level adapter, wherein a $1\times1$ convolution is added to each residual block in series. Only the task-specific adapter layers are trained in each of these models, while the $W_s$ weights are frozen to Imagenet pre-training. RCM-NFI applies normalized feature fusion to the output of RCM layers in the RCM-I model (given as RCM-NFF in~\cite{kanakis2020reparameterizing}). This model does not perform well in our setting. The RAP is a plug-and-play residual adapter that can easily be plugged into existing segmentation models. The RAS and RCM are series adapters and need to be included when ResNet is pre-trained on Imagenet for best performance. We find that the RAP adapter is better suited when fine-tuning the shared weights $W_s$ (please see supplementary material). 

\noindent \textbf{Latent space visualization.}
Figure~\ref{fig:tsne} shows a t-SNE~\cite{van2008visualizing} visualization of the features extracted from the last layer of the encoder $\mathcal{F}( . )$. We show the latent space of features extracted from a Cityscapes sample image before and after incrementally learning over the next domain BDD (CS $\rightarrow$ BDD setting). In (a), we train a single-task baseline on CS in step 1 and fine-tune it on BDD in step 2. In (b), we train \textit{our} model on CS in step 1 and incrementally learn on BDD in step 2. The latent space of CS gets significantly distorted after fine-tuning the model on BDD (see (a) Step 2). On the other hand, our model can preserve the learned feature representation even after adding BDD to the model. While the smaller classes (with fewer pixels) had distinct separation in step 1, inter-cluster separation has suffered in step 2 for the fine-tuning model. It can be observed that classes like rider, motorcycle, traffic light, bicycle and truck have moved in a small space towards the center and are indistinguishable, causing confusion in predicting these classes. Our model in step 2 is able to preserve distinct clusters and maintain inter-class separation on the old domain CS.

\vspace{-6pt}
\section{Conclusion}
\vspace{-6pt}
We define the problem of multi-domain incremental semantic segmentation and present a parameter isolation-based dynamic architecture that leads to a significant improvement over the baselines. Our model allows domain-specific paths for different domains while having a large degree of parameter sharing (78.83\%) in a universal model. We compare two scenarios of fully overlapping and partially overlapping label spaces to understand the challenges involved in multi-domain incremental semantic segmentation. From the CS $\rightarrow$ BDD and CS $\rightarrow$ IDD cases, we infer that: (i) our approach works equally well in mitigating forgetting in the two scenarios; (ii) if the labels spaces are aligned, a forward transfer can occur; (iii) misalignment of label spaces is likely to cause some domain interference on the new domain, although our method provides promising performance across the domains. We demonstrate through visualizations how the proposed method maintains the latent space of classes across domains. This enables learning novel domains while preserving the representations of the previous domains at the same time. 

\noindent \textbf{Acknowledgments.} This work was partly funded by IHub-Data at IIIT Hyderabad, and DST (IMPRINT program).

{\small
\bibliographystyle{ieee_fullname}
\bibliography{egbib}

\begin{thebibliography}{10}\itemsep=-1pt

\bibitem{aljundi2018memory}
Rahaf Aljundi, Francesca Babiloni, Mohamed Elhoseiny, Marcus Rohrbach, and
  Tinne Tuytelaars.
\newblock Memory {Aware Synapses: L}earning what (not) to forget.
\newblock In {\em Proceedings of the European Conference on Computer Vision
  (ECCV)}, pages 139--154, 2018.

\bibitem{aljundi2017expert}
Rahaf Aljundi, Punarjay Chakravarty, and Tinne Tuytelaars.
\newblock Expert {G}ate: {Lifelong Learning with a Network of E}xperts.
\newblock In {\em Proceedings of the IEEE/CVF Conference on Computer Vision and
  Pattern Recognition (CVPR)}, pages 3366--3375, 2017.

\bibitem{bulat2020incremental}
Adrian Bulat, Jean Kossaifi, Georgios Tzimiropoulos, and Maja Pantic.
\newblock Incremental {Multi-Domain Learning with Network Latent Tensor
  F}actorization.
\newblock In {\em Proceedings of the AAAI Conference on Artificial
  Intelligence}, volume~34, pages 10470--10477, 2020.

\bibitem{castro2018end}
Francisco~M Castro, Manuel~J Mar{\'\i}n-Jim{\'e}nez, Nicol{\'a}s Guil, Cordelia
  Schmid, and Karteek Alahari.
\newblock End-to-{End Incremental L}earning.
\newblock In {\em Proceedings of the European conference on Computer Vision
  (ECCV)}, pages 233--248, 2018.

\bibitem{cermelli2020modeling}
Fabio Cermelli, Massimiliano Mancini, Samuel~Rota Bulo, Elisa Ricci, and
  Barbara Caputo.
\newblock Modeling the {Background for Incremental Learning in Semantic
  S}egmentation.
\newblock In {\em Proceedings of the IEEE/CVF Conference on Computer Vision and
  Pattern Recognition (CVPR)}, pages 9233--9242, 2020.

\bibitem{chaudhry2018riemannian}
Arslan Chaudhry, Puneet~K Dokania, Thalaiyasingam Ajanthan, and Philip~HS Torr.
\newblock Riemannian {Walk for Incremental Learning: Understanding Forgetting
  and I}ntransigence.
\newblock In {\em Proceedings of the European Conference on Computer Vision
  (ECCV)}, pages 532--547, 2018.

\bibitem{cordts2016cityscapes}
Marius Cordts, Mohamed Omran, Sebastian Ramos, Timo Rehfeld, Markus Enzweiler,
  Rodrigo Benenson, Uwe Franke, Stefan Roth, and Bernt Schiele.
\newblock The {Cityscapes Dataset for Semantic Urban Scene U}nderstanding.
\newblock In {\em Proceedings of the IEEE/CVF conference on Computer Vision and
  Pattern Recognition (CVPR)}, pages 3213--3223, 2016.

\bibitem{de2019continual}
Matthias De~Lange, Rahaf Aljundi, Marc Masana, Sarah Parisot, Xu Jia, Ales
  Leonardis, Gregory Slabaugh, and Tinne Tuytelaars.
\newblock Continual learning: {A} comparative study on how to defy forgetting
  in classification tasks.
\newblock {\em arXiv preprint arXiv:1909.08383}, 2(6), 2019.

\bibitem{delange2021continual}
Matthias Delange, Rahaf Aljundi, Marc Masana, Sarah Parisot, Xu Jia, Ales
  Leonardis, Greg Slabaugh, and Tinne Tuytelaars.
\newblock A continual learning survey: {D}efying forgetting in classification
  tasks.
\newblock {\em IEEE Transactions on Pattern Analysis and Machine Intelligence
  (TPAMI)}, 2021.

\bibitem{deng2009imagenet}
Jia Deng, Wei Dong, Richard Socher, Li-Jia Li, Kai Li, and Li Fei-Fei.
\newblock Image{Net: A Large-Scale Hierarchical Image D}atabase.
\newblock In {\em Proceedings of the IEEE/CVF conference on Computer Vision and
  Pattern Recognition (CVPR)}, pages 248--255. IEEE, 2009.

\bibitem{dhar2019learning}
Prithviraj Dhar, Rajat~Vikram Singh, Kuan-Chuan Peng, Ziyan Wu, and Rama
  Chellappa.
\newblock Learning without {M}emorizing.
\newblock In {\em Proceedings of the IEEE/CVF Conference on Computer Vision and
  Pattern Recognition (CVPR)}, pages 5138--5146, 2019.

\bibitem{douillard2020plop}
Arthur Douillard, Yifu Chen, Arnaud Dapogny, and Matthieu Cord.
\newblock P{LOP: Learning without Forgetting for Continual Semantic
  S}egmentation.
\newblock In {\em Proceedings of the IEEE/CVF Conference on Computer Vision and
  Pattern Recognition (CVPR)}, pages 4040--4050, 2021.

\bibitem{ebrahimi2020adversarial}
Sayna Ebrahimi, Franziska Meier, Roberto Calandra, Trevor Darrell, and Marcus
  Rohrbach.
\newblock Adversarial {Continual L}earning.
\newblock {\em arXiv preprint arXiv:2003.09553}, 1, 2020.

\bibitem{guo2019depthwise}
Yunhui Guo, Yandong Li, Liqiang Wang, and Tajana Rosing.
\newblock Depthwise {Convolution Is All You Need for Learning Multiple Visual
  D}omains.
\newblock In {\em Proceedings of the AAAI Conference on Artificial
  Intelligence}, volume~33, pages 8368--8375, 2019.

\bibitem{he2016deep}
Kaiming He, Xiangyu Zhang, Shaoqing Ren, and Jian Sun.
\newblock Deep {Residual Learning for Image R}ecognition.
\newblock In {\em Proceedings of the IEEE/CVF conference on Computer Vision and
  Pattern Recognition (CVPR)}, pages 770--778, 2016.

\bibitem{hinton2015distilling}
Geoffrey Hinton, Oriol Vinyals, and Jeff Dean.
\newblock Distilling the {Knowledge in a Neural N}etwork.
\newblock {\em stat}, 1050:9, 2015.

\bibitem{hou2019learning}
Saihui Hou, Xinyu Pan, Chen~Change Loy, Zilei Wang, and Dahua Lin.
\newblock Learning a {Unified Classifier Incrementally via R}ebalancing.
\newblock In {\em Proceedings of the IEEE/CVF Conference on Computer Vision and
  Pattern Recognition (CVPR)}, pages 831--839, 2019.

\bibitem{kalluri2019universal}
Tarun Kalluri, Girish Varma, Manmohan Chandraker, and CV Jawahar.
\newblock Universal {Semi-Supervised Semantic S}egmentation.
\newblock In {\em Proceedings of the IEEE/CVF International Conference on
  Computer Vision (ICCV)}, pages 5259--5270, 2019.

\bibitem{kanakis2020reparameterizing}
Menelaos Kanakis, David Bruggemann, Suman Saha, Stamatios Georgoulis, Anton
  Obukhov, and Luc Van~Gool.
\newblock Reparameterizing {Convolutions for Incremental Multi-Task Learning
  without Task I}nterference.
\newblock In {\em Proceedings of the European Conference on Computer Vision
  (ECCV)}, pages 689--707. Springer, 2020.

\bibitem{kirkpatrick2017overcoming}
James Kirkpatrick, Razvan Pascanu, Neil Rabinowitz, Joel Veness, Guillaume
  Desjardins, Andrei~A Rusu, Kieran Milan, John Quan, Tiago Ramalho, Agnieszka
  Grabska-Barwinska, et~al.
\newblock Overcoming catastrophic forgetting in neural networks.
\newblock {\em Proceedings of the National Academy of Sciences},
  114(13):3521--3526, 2017.

\bibitem{klingner2020class}
Marvin Klingner, Andreas B{\"a}r, Philipp Donn, and Tim Fingscheidt.
\newblock Class-{Incremental Learning for Semantic Segmentation Re-Using
  Neither Old Data Nor Old L}abels.
\newblock In {\em 2020 IEEE 23rd International Conference on Intelligent
  Transportation Systems (ITSC)}, pages 1--8. IEEE, 2020.

\bibitem{kothandaraman2021domain}
Divya Kothandaraman, Athira~M Nambiar, and Anurag Mittal.
\newblock Domain {Adaptive Knowledge Distillation for Driving Scene Semantic
  S}egmentation.
\newblock In {\em Proceedings of the IEEE Winter Conference on Applications of
  Computer Vision (WACV Workshops)}, pages 134--143, 2021.

\bibitem{kundu2020class}
Jogendra~Nath Kundu, Rahul~Mysore Venkatesh, Naveen Venkat, Ambareesh Revanur,
  and R~Venkatesh Babu.
\newblock Class-{Incremental Domain A}daptation.
\newblock In {\em Proceedings of the European Conference on Computer Vision
  (ECCV)}, pages 53--69, 2020.

\bibitem{li2017learning}
Zhizhong Li and Derek Hoiem.
\newblock Learning without {F}orgetting.
\newblock {\em IEEE Transactions on Pattern Analysis and Machine Intelligence
  (TPAMI)}, 40(12):2935--2947, 2017.

\bibitem{liu2020multi}
Xialei Liu, Hao Yang, Avinash Ravichandran, Rahul Bhotika, and Stefano Soatto.
\newblock Multi-{Task Incremental Learning for Object D}etection.
\newblock {\em arXiv e-prints}, pages arXiv--2002, 2020.

\bibitem{mallya2018piggyback}
Arun Mallya, Dillon Davis, and Svetlana Lazebnik.
\newblock Piggyback: {Adapting a Single Network to Multiple Tasks by Learning
  to Mask W}eights.
\newblock In {\em Proceedings of the European Conference on Computer Vision
  (ECCV)}, pages 67--82, 2018.

\bibitem{mallya2018packnet}
Arun Mallya and Svetlana Lazebnik.
\newblock Packnet: Adding multiple tasks to a single network by iterative
  pruning.
\newblock In {\em Proceedings of the IEEE/CVF Conference on Computer Vision and
  Pattern Recognition (CVPR)}, pages 7765--7773, 2018.

\bibitem{mancini2018adding}
Massimiliano Mancini, Elisa Ricci, Barbara Caputo, and Samuel Rota~Bulo.
\newblock Adding {New Tasks to a Single Network with Weight Transformations
  using Binary M}asks.
\newblock In {\em Proceedings of the European Conference on Computer Vision
  (ECCV) Workshops}, pages 180--189, 2018.

\bibitem{mccloskey1989catastrophic}
Michael McCloskey and Neal~J Cohen.
\newblock Catastrophic {Interference in Connectionist Networks: The Sequential
  Learning P}roblem.
\newblock In {\em Psychology of learning and motivation}, volume~24, pages
  109--165. Elsevier, 1989.

\bibitem{mei2020instance}
Ke Mei, Chuang Zhu, Jiaqi Zou, and Shanghang Zhang.
\newblock Instance {Adaptive Self-Training for Unsupervised Domain A}daptation.
\newblock {\em arXiv preprint arXiv:2008.12197}, 2020.

\bibitem{mermillod2013stability}
Martial Mermillod, Aur{\'e}lia Bugaiska, and Patrick Bonin.
\newblock The stability-plasticity dilemma: Investigating the continuum from
  catastrophic forgetting to age-limited learning effects.
\newblock {\em Frontiers in psychology}, 4:504, 2013.

\bibitem{michieli2019incremental}
Umberto Michieli and Pietro Zanuttigh.
\newblock Incremental {Learning Techniques for Semantic S}egmentation.
\newblock In {\em Proceedings of the IEEE/CVF International Conference on
  Computer Vision Workshops (ICCVW)}, pages 3205--3212, 2019.

\bibitem{michieli2021continual}
Umberto Michieli and Pietro Zanuttigh.
\newblock Continual {Semantic Segmentation via Repulsion-Attraction of Sparse
  and Disentangled Latent R}epresentations.
\newblock In {\em Proceedings of the IEEE/CVF Conference on Computer Vision and
  Pattern Recognition (CVPR)}, pages 1114--1124, 2021.

\bibitem{ostapenko2019learning}
Oleksiy Ostapenko, Mihai Puscas, Tassilo Klein, Patrick Jahnichen, and Moin
  Nabi.
\newblock Learning to {Remember: A Synaptic Plasticity Driven Framework for
  Continual L}earning.
\newblock In {\em Proceedings of the IEEE/CVF Conference on Computer Vision and
  Pattern Recognition (CVPR)}, pages 11321--11329, 2019.

\bibitem{parisi2019continual}
German~I Parisi, Ronald Kemker, Jose~L Part, Christopher Kanan, and Stefan
  Wermter.
\newblock Continual {Lifelong Learning with Neural Networks: A R}eview.
\newblock {\em Neural Networks}, 113:54--71, 2019.

\bibitem{rebuffi2017learning}
S-A Rebuffi, H. Bilen, and A. Vedaldi.
\newblock Learning multiple visual domains with residual adapters.
\newblock In {\em Advances in Neural Information Processing Systems (NIPS)},
  2017.

\bibitem{rebuffi2018efficient}
Sylvestre-Alvise Rebuffi, Hakan Bilen, and Andrea Vedaldi.
\newblock Efficient parametrization of multi-domain deep neural networks.
\newblock In {\em Proceedings of the IEEE/CVF Conference on Computer Vision and
  Pattern Recognition (CVPR)}, pages 8119--8127, 2018.

\bibitem{rebuffi2017icarl}
Sylvestre-Alvise Rebuffi, Alexander Kolesnikov, Georg Sperl, and Christoph~H
  Lampert.
\newblock i{CaRL: Incremental Classifier and Representation L}earning.
\newblock In {\em Proceedings of the IEEE/CVF conference on Computer Vision and
  Pattern Recognition (CVPR)}, pages 2001--2010, 2017.

\bibitem{romera2017erfnet}
Eduardo Romera, Jos{\'e}~M Alvarez, Luis~M Bergasa, and Roberto Arroyo.
\newblock E{RFNet: Efficient Residual Factorized ConvNet for Real-Time Semantic
  S}egmentation.
\newblock {\em IEEE Transactions on Intelligent Transportation Systems},
  19(1):263--272, 2017.

\bibitem{rosenfeld2018incremental}
Amir Rosenfeld and John~K Tsotsos.
\newblock Incremental {Learning Through Deep A}daptation.
\newblock {\em IEEE Transactions on Pattern Analysis and Machine Intelligence
  (TPAMI)}, 42(3):651--663, 2018.

\bibitem{rusu2016progressive}
Andrei~A Rusu, Neil~C Rabinowitz, Guillaume Desjardins, Hubert Soyer, James
  Kirkpatrick, Koray Kavukcuoglu, Razvan Pascanu, and Raia Hadsell.
\newblock Progressive {Neural N}etworks.
\newblock {\em arXiv preprint arXiv:1606.04671}, 2016.

\bibitem{shin2017continual}
Hanul Shin, Jung~Kwon Lee, Jaehong Kim, and Jiwon Kim.
\newblock Continual {Learning with Deep Generative R}eplay.
\newblock In {\em Advances in Neural Information Processing Systems (NIPS)},
  2017.

\bibitem{singh2020calibrating}
Pravendra Singh, Vinay~Kumar Verma, Pratik Mazumder, Lawrence Carin, and Piyush
  Rai.
\newblock Calibrating {CNNs for Lifelong L}earning.
\newblock {\em Advances in Neural Information Processing Systems (NIPS)}, 33,
  2020.

\bibitem{van2008visualizing}
Laurens Van~der Maaten and Geoffrey Hinton.
\newblock Visualizing {Data using t-SNE}.
\newblock {\em Journal of machine learning research}, 9(11), 2008.

\bibitem{varma2019idd}
Girish Varma, Anbumani Subramanian, Anoop Namboodiri, Manmohan Chandraker, and
  CV Jawahar.
\newblock I{DD: A Dataset for Exploring Problems of Autonomous Navigation in
  Unconstrained E}nvironments.
\newblock In {\em Proceedings of the IEEE Winter Conference on Applications of
  Computer Vision (WACV)}, pages 1743--1751. IEEE, 2019.

\bibitem{vu2019advent}
Tuan-Hung Vu, Himalaya Jain, Maxime Bucher, Matthieu Cord, and Patrick
  P{\'e}rez.
\newblock A{DVENT: Adversarial Entropy Minimization for Domain Adaptation in
  Semantic S}egmentation.
\newblock In {\em Proceedings of the IEEE/CVF Conference on Computer Vision and
  Pattern Recognition (CVPR)}, pages 2517--2526, 2019.

\bibitem{wang2020classes}
Haoran Wang, Tong Shen, Wei Zhang, Ling-Yu Duan, and Tao Mei.
\newblock Classes {Matter: A Fine-grained Adversarial Approach to Cross-domain
  Semantic S}egmentation.
\newblock In {\em Proceedings of the European Conference on Computer Vision
  (ECCV)}, pages 642--659. Springer, 2020.

\bibitem{wang2021cross}
Li Wang, Dong Li, Yousong Zhu, Lu Tian, and Yi Shan.
\newblock Cross-{Dataset Collaborative Learning for Semantic S}egmentation.
\newblock {\em arXiv preprint arXiv:2103.11351}, 2021.

\bibitem{wu2018memory}
Chenshen Wu, Luis Herranz, Xialei Liu, Yaxing Wang, Joost van~de Weijer, and
  Bogdan Raducanu.
\newblock Memory {Replay GAN}s: learning to generate images from new categories
  without forgetting.
\newblock In {\em Conference on Neural Information Processing Systems (NIPS)},
  2018.

\bibitem{wu2019large}
Yue Wu, Yinpeng Chen, Lijuan Wang, Yuancheng Ye, Zicheng Liu, Yandong Guo, and
  Yun Fu.
\newblock Large {Scale Incremental L}earning.
\newblock In {\em Proceedings of the IEEE/CVF Conference on Computer Vision and
  Pattern Recognition (CVPR)}, pages 374--382, 2019.

\bibitem{wu2019ace}
Zuxuan Wu, Xin Wang, Joseph~E Gonzalez, Tom Goldstein, and Larry~S Davis.
\newblock A{CE: Adapting to Changing Environments for Semantic S}egmentation.
\newblock In {\em Proceedings of the IEEE/CVF International Conference on
  Computer Vision (ICCV)}, pages 2121--2130, 2019.

\bibitem{wulfmeier2018incremental}
Markus Wulfmeier, Alex Bewley, and Ingmar Posner.
\newblock Incremental {Adversarial Domain Adaptation for Continually Changing
  E}nvironments.
\newblock In {\em 2018 IEEE International Conference on Robotics and Automation
  (ICRA)}, pages 4489--4495. IEEE, 2018.

\bibitem{yang2020fda}
Yanchao Yang and Stefano Soatto.
\newblock F{DA: Fourier Domain Adaptation for Semantic S}egmentation.
\newblock In {\em Proceedings of the IEEE/CVF Conference on Computer Vision and
  Pattern Recognition (CVPR)}, pages 4085--4095, 2020.

\bibitem{yoon2017lifelong}
Jaehong Yoon, Eunho Yang, Jungtae Lee, and Sung~Ju Hwang.
\newblock Lifelong {Learning with Dynamically Expandable N}etworks.
\newblock In {\em Sixth International Conference on Learning Representations}.
  ICLR, 2018.

\bibitem{yu2020bdd100k}
Fisher Yu, Haofeng Chen, Xin Wang, Wenqi Xian, Yingying Chen, Fangchen Liu,
  Vashisht Madhavan, and Trevor Darrell.
\newblock B{DD100K: A Diverse Driving Dataset for Heterogeneous Multitask
  L}earning.
\newblock In {\em Proceedings of the IEEE/CVF conference on Computer Vision and
  Pattern Recognition (CVPR)}, pages 2636--2645, 2020.

\bibitem{zenke2017continual}
Friedemann Zenke, Ben Poole, and Surya Ganguli.
\newblock Continual {Learning Through Synaptic I}ntelligence.
\newblock In {\em International Conference on Machine Learning}, pages
  3987--3995. PMLR, 2017.

\bibitem{zhao2019multi}
Sicheng Zhao, Bo Li, Xiangyu Yue, Yang Gu, Pengfei Xu, Runbo Tan, Hu, Hua Chai,
  and Kurt Keutzer.
\newblock Multi-source {Domain Adaptation for Semantic S}egmentation.
\newblock In {\em Advances in Neural Information Processing Systems (NIPS)},
  2019.

\end{thebibliography}
}

\end{document}